%% file: egpaper_final.tex
\documentclass[10pt,twocolumn,letterpaper]{article}

\usepackage{cvpr}
\usepackage{times}
\usepackage{epsfig}
\usepackage{graphicx}
\usepackage{amsmath}
\usepackage{amssymb}
\usepackage{comment,float}

\usepackage{setspace}

\usepackage{epsfig}
\usepackage{graphicx}
\usepackage{amsmath}
\usepackage{amssymb}
\usepackage{algorithm2e}

\usepackage[pagebackref=true,breaklinks=true,letterpaper=true,colorlinks,bookmarks=false]{hyperref}

\cvprfinalcopy 

\def\cvprPaperID{39} 

\begin{document}

\title{3D Face Hallucination from a Single Depth Frame}

\author{Shu Liang, Ira Kemelmacher-Shlizerman, Linda G. Shapiro\\
University of Washington\\
185 Stevens Way, WA 98105\\
{\tt\small \{liangshu,kemelmi,shapiro\}@cs.washington.edu}
}

\maketitle


\input{abstract}
\input{intro}

\input{method}

\input{sim_eval}

\input{result}

\input{conclusion}

\section*{Acknowledgements}
This work was supported by the National Institute of Dental and Craniofacial Research under Grant No. U01-DE02005. Morphometric data from normal faces were obtained from FaceBase (www.facebase.org), and were generated by projects U01DE020078 and U01DE020054. The FaceBase Data Management Hub (U01DE020057) and the FaceBase Consortium are funded by the National Institute of Dental and Craniofacial Research.

{\small
\bibliographystyle{ieee}
\bibliography{egpaper_final,eccv14}
}

\end{document}

%% file: abstract.tex
\begin{abstract}
We present an algorithm that takes a single frame of a person's face from a depth camera, e.g., Kinect, and produces a high-resolution 3D mesh of the input face. We leverage a  dataset of  3D face meshes of $1204$ distinct  individuals ranging from age 3 to 40, captured in a neutral expression. We divide the input depth frame into semantically significant regions (eyes, nose, mouth, cheeks) and search the database for the best matching  shape per region. We further combine the input depth frame with the matched database shapes into a single mesh that results in a high-resolution shape of the input person.  Our system is fully automatic and uses only depth data for matching, making it invariant to imaging conditions.  We evaluate our results using ground truth shapes,  as well as compare to state-of-the-art shape estimation methods.  We demonstrate the robustness of our local matching approach with high-quality reconstruction of faces that fall \textit{outside} of the dataset span, e.g., faces older than 40 years old, facial expressions, and different ethnicities.  
\end{abstract}

%% file: intro.tex
\section{Introduction}

Acquiring high-detail 3D face meshes is  challenging due to the highly non-rigid nature of human faces. 
High-detail reconstruction methods currently require the subject to come to a lab  equipped with a calibrated set of cameras and/or lights, e.g., multi-view stereo approaches \cite{beeler2010high,beeler2011high,bradley2010high}, structured light \cite{zhang2007spacetime}, and light stages \cite{alexander2013digital,alexander2009digital,ghosh2011multiview}. For many applications, however, we would like to enable scanning capabilities \textit{anywhere}. Indeed, the proliferation of depth cameras can potentially allow  shape capturing even in the comfort's of one's home. For example,  KinectFusion \cite{newcombe2011kinectfusion}  allows high quality  capture by moving a depth camera around the subject. It requires, however,  the subject to stay still (with the same facial expression) during the capturing session. 

In this paper, we demonstrate that  high quality shape can be captured from a \textit{single} depth view.  Most single view methods use as input only the intensity or color information and thus prone to gauge and bas-relief ambiguities \cite{kemelmacher20113d}. Recently, \cite{weise2011realtime,bouaziz2013online} have shown  impressive face tracking and re-targeting results from Kinect input. The reconstructed shape, however, typically lacks details, since it is assumed to be in a linear span of the scans used to create a morphable model \cite{blanz1999morphable,anguelov2005scape}. Instead in this paper, we choose a single best database mesh per facial part, and then merge the individual parts, rather than assuming that the shape is spanned by a database. This enables high-detail shape reconstructions.  In Fig. \ref{fig:teaser} we show example results that were automatically produced by our algorithm.  

\begin{figure}
\centering
\includegraphics[width=.5\textwidth]{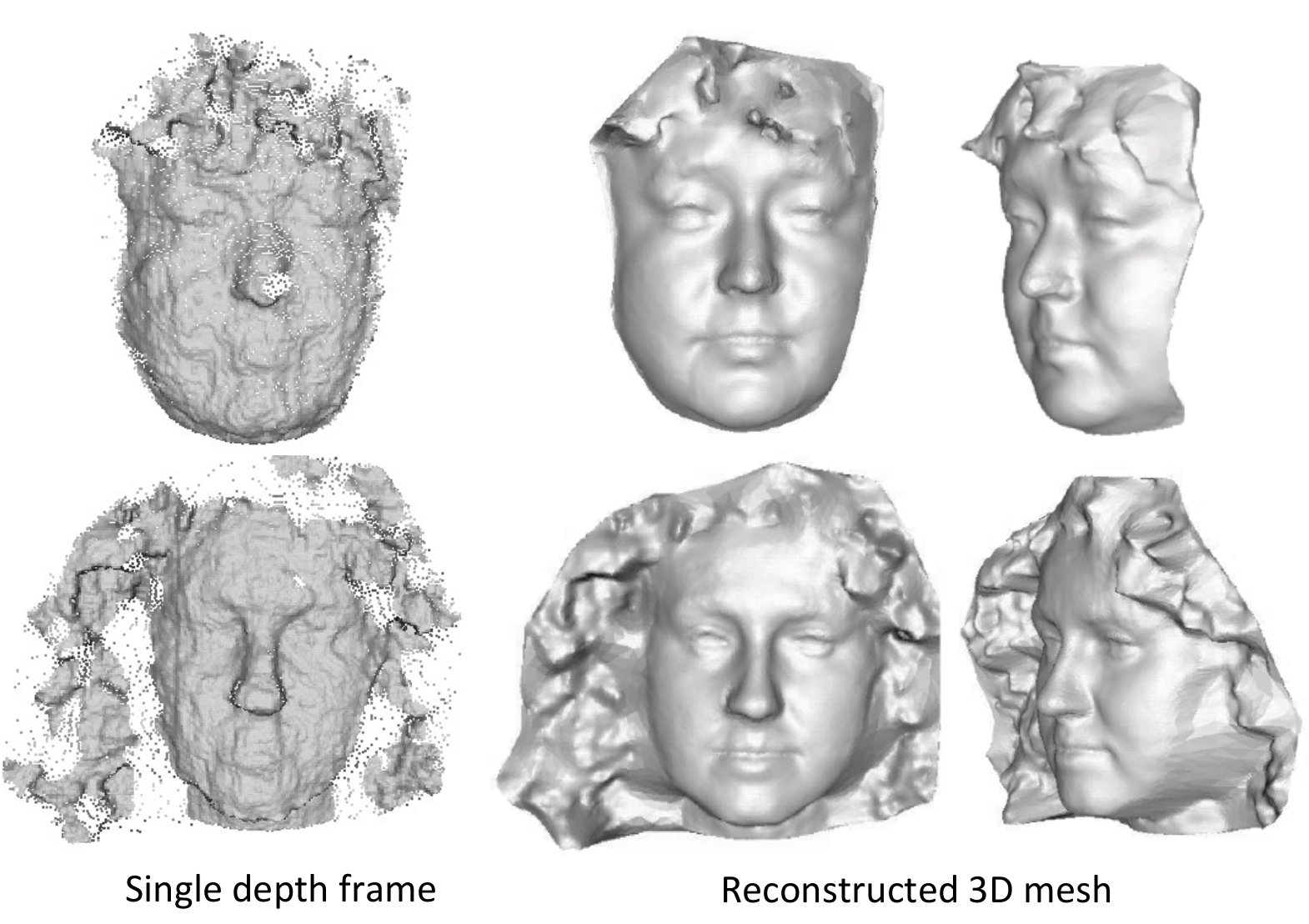}
\caption{Our approach takes as input a \textit{single} depth frame of a person's face and outputs a high-resolution 3D mesh of the face completely automatically.}
\end{figure}\label{fig:teaser}

The key idea of this work is that while a single depth frame of a person's face is extremely noisy and low resolution, it still encodes metric information about the person's underlying facial features.  Our approach is to leverage a large dataset of 3D face scans (1204 meshes of distinct Caucasian individuals, with age ranging from 3 to 40) for \textit{hallucination} of a new 3D shape. We are inspired by texture synthesis approaches that leverage a large number of photos to fill in missing parts in a new photo \cite{hays2007scene}. However, instead of working with photos, we propose an approach that finds similarities between a  depth image and high-resolution 3D scans. Related to our work are also shape matching approaches such as \cite{pokrass2013partial,kovnatsky2013coupled}, our goal is however different since rather than searching for corresponding semantic parts we search for best matches for a particular part.  Specifically, we match small parts from the depth frame to  parts of the dataset faces, copy the matched parts from the corresponding dataset meshes and finally combine them together. This approach works remarkably well and can even reconstruct shapes of people who fall outside of the dataset span, such as, for people of older age and Asian ethnicity.  

\begin{figure*}
\begin{center}
  \includegraphics[width=1\textwidth]{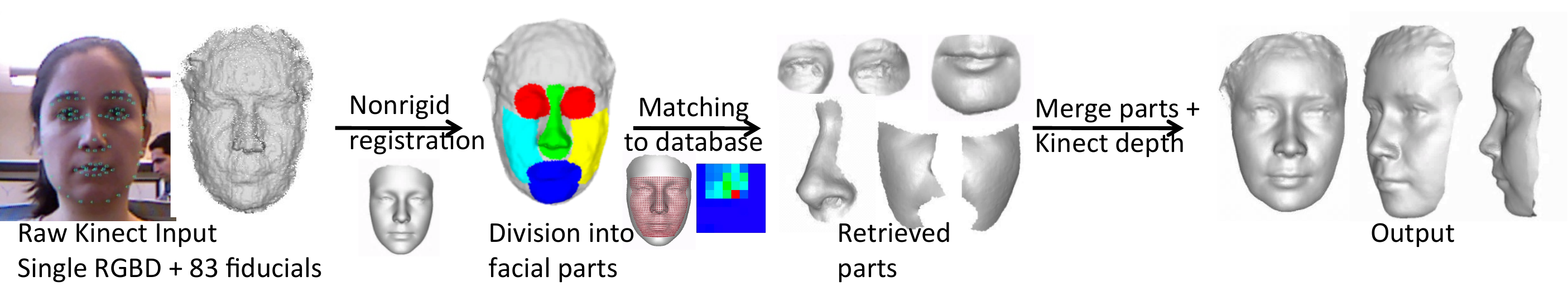}
\end{center}
     \caption{Overview of our approach.}\label{fig:pipeline}
\end{figure*}

The paper is organized as follows. We begin by describing our full reconstruction approach, which we call \textit{3D Hallucination}, in Sec.~\ref{sec:method}.  In Section~\ref{sec:sim_eval} we define and evaluate our distance function that was used to match a Kinect frame to the dataset. In Section~\ref{sec:results} we describe the dataset, and compare to ground-truth and related methods.

%% file: method.tex
\section{3D Hallucination}\label{sec:method}

\begin{figure}
\begin{center}
   \includegraphics[width=0.9\linewidth]{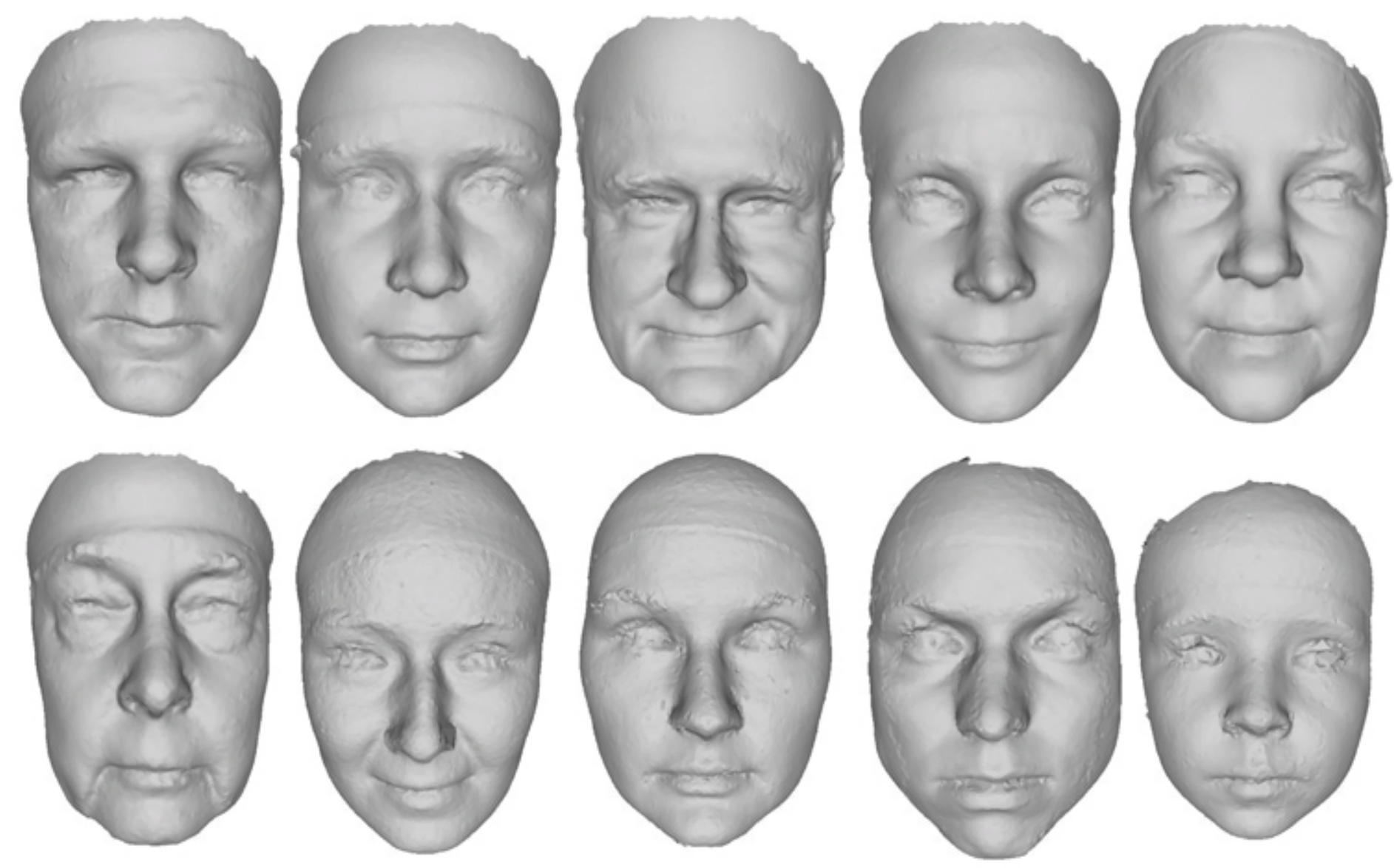}
\end{center}
   \caption{Example high-resolution face meshes. The database includes meshes (no texture) of 652 females and 552 males, ages 3 to 40, captured in a neutral expression. }
\label{fig:database}
\end{figure}

In this section, we describe our complete approach that takes as input a single RGBD frame of a person's face and outputs a high-resolution 3D mesh of the input face.  We are given a large dataset of  high-resolution 3D face meshes (just the mesh, without texture), captured in a neutral expression. Examples of high-resolution meshes are shown in Fig.~\ref{fig:database}. All the meshes in the dataset have been put into dense correspondence using \cite{allen}. Further, the aligned database meshes are averaged to produce the generic mesh $G$. Finally, we define five facial areas on $G$ and, using the dense correspondence, propagate the areas to  the database meshes. 

Our approach  is as follows. We first align the input RGBD frame to the generic mesh $G$. Then the input depth is divided into five facial parts via the alignment, and each facial part is matched independently to the dataset resulting in five high-resolution meshes. Finally, the matched meshes are combined with the input into a single mesh to produce the output.  Fig.~\ref{fig:pipeline} illustrates all these steps. Below, we describe each of the steps in detail.

\subsection{Aligning a single depth frame to the database}
\label{sec:align}

Given a single RGBD frame of a person's face in neutral facial expression, we first detect the face and 83 fiducial points. Any facial landmark detection method can be applied on RGB\cite{fan2014learning, cao2014face} or depth image\cite{fanelli2013random}. We use the software of Face++ \cite{face++}. Out of the 83 points, 19 are on the silhouette of the face, and the rest are on the internal part of the face. We use the  internal facial points for  rigid pose alignment via Procrustes analysis \cite{gower1975generalized} and then all 83 points for dense alignment to the generic mesh $G$ \cite{allen}. We obtain point-to-point correspondence between the depth frame and the generic shape, producing a deformed generic mesh $G'$ which minimizes the difference to the depth frame. With the 83 points, all the faces in our data set are warped using \cite{pighin2006synthesizing} so that their global shapes are deformed to match the input depth image better. We define five facial parts on the input depth image based on the correspondence to the generic mesh. The five facial parts correspond to eyes, nose, mouth, left cheek, and right cheek as illustrated in Fig.~\ref{fig:pipeline}.

\subsection{Part-based matching to the database}
\label{sec:match}
The next step is to match each of the five facial parts in the input frame to the database.  Prior to the matching process, we apply a curvature flow smoothing method \cite{desbrun1999implicit} that preserves the  low-frequency shape while smoothing out the noise.  

Each of the five facial parts is then matched to the database using our distance function. The distance is a weighted combination of pseudo-landmarks and  histograms of  azimuth and elevation components of the surface normals, following \cite{mercan2013use,atmosukarto2010use}. 
The distance function is described in detail in Sec.~\ref{sec:sim_eval}. The matching process results in five high-resolution meshes that are retrieved from the database.  Each mesh matches to a different part of the input face.

\subsection{Merging the matches}
\label{sec:merge}

Once we get the five matches, the vertex normals are copied to replace the original normals of deformed generic shape $G'$, part by part. Our query mesh can have hair while the high-resolution 3D head models do not. For each vertex $V$ in the face region, using the nearest triangle ${\triangle}ABC$ in $G'$, the normal vector of $V$ can be interpolated as the weighted combination of the normal directions of ${\triangle}ABV$, ${\triangle}VBC$ and ${\triangle}VCA$. For the hair region, the original normals are kept. After we compute new normals for each vertex in the face region, we fuse the depth from the Kinect frame and the new normals together using the method of \cite{nehab2005efficiently}. Then fine details on the facial part are transferred to the input face, but the hair style is kept.

\subsection{Facial expressions}

The above process produces a high-resolution mesh of the input face from a single noisy Kinect frame.  While the focus of this work is on neutral faces, we further show that it is possible to produce high-resolution meshes of \textit{facial expressions} using the same approach.  It is  challenging to acquire a database of high-resolution meshes of many distinct individuals making a large number of facial expressions.  Instead, we show that given a single RGBD frame of a person in neutral expression and another frame that captures a facial expression, our approach can output a high-resolution expression mesh. 

Specifically, we retrieve five matches from the database using the neutral input as described in \ref{sec:align} and \ref{sec:match}, and then include the expression depth frame in the merging process. Each of the five database meshes are deformed towards the expression frame as in \ref{sec:align}, and then we execute exactly the same merging process as in \ref{sec:merge}.

%% file: sim_eval.tex
\section{Similarity function}\label{sec:sim_eval}

In this section we describe our similarity function. It is  used to match each of the five facial parts of the input depth frame to the corresponding parts of the database meshes. The similarity function is a weighted combination of \textit{pseudo-landmarks} and \textit{histograms of azimuth-elevation} components of the surface normals.

\textbf{Pseudo-landmarks.} To obtain pseudo-landmarks we sample the  Kinect shape and each of the database meshes (which are at that stage in dense correspondence) following \cite{mercan2013use}.
First, two anatomical landmarks (the sellion and chin tip), are computed and two base horizontal planes are computed through these points. Then, $m$ parallel planes are computed between the two base planes, each sampled by $n$ points.  We chose $m=33$ and $n=35$ for a total of $1,225$ points, resulting in $19,033$ vertices. Additional details are described in the evaluation part below. Once pseudo-landmarks are estimated, the distance per database mesh $j$ is defined as  
\begin{equation}
D_\text{pts}^j=\sum_{i=1}^{(m+2)n}||P^j_{i}-P^\text{input}_i||^2
\end{equation}
where $P_{i}^{*}$ is an xyz-coordinate of a pseudo-landmark. 

\textbf{Histograms of azimuth-elevation.} We also compute distances between surface normals, as follows.  Given the surface normal $\vec{n}=(n_x,n_y,n_z)$ at a point, the azimuth angle $\theta$ is defined as the angle between the positive $x$-axis and the projection of $\vec{n}$ to the $xy$ plane. The elevation angle $\phi$ is the angle between the $x$-axis and $\vec{n}$:
\begin{equation}
\theta=\arctan(\frac{n_z}{n_x}) ,         \phi=\arctan(\frac{n_y}{\sqrt{({n_x}^2+{n_z}^2)}})
\end{equation}
with $\theta{\in}[-\pi,\pi]$, $\phi{\in}[-\frac{\pi}{2},\frac{\pi}{2}]$.
Histograms are useful to determine the ``flatness'' and the dominant orientation of a surface patch.  We calculate a $32{\times}32$ histogram for each facial component, and define the distance as the $\chi^2$-distance between the histograms
\begin{equation}
D_\text{normals}=\chi^2(H^j,H^\text{input}).
\end{equation}

\textbf{Combined distance.}
The combined distance for a single facial part  is then defined as
\begin{equation}
D=D_\text{pts}+\alpha D_\text{normals}
\end{equation}
The parameter $\alpha$ is chosen per facial part  according to our evaluation experiment in Section \ref{sec:sim_evaluation}. The cheek area typically has less variation in  surface normals across points and thus has a small $\alpha=1$; the mouth has higher normal variation and thus $\alpha$  will be larger ($\alpha=10$). We chose $\alpha=4$ for the eye area and $\alpha=2$ for the nose area.

%% file: result.tex
\section{Experiments}\label{sec:results}
Below we describe the details of our data, our implementation, and our results.

\subsection{Implementation and data details}\label{sec:data}

We used a Microsoft Kinect to capture the inputs in resolution $640\times 480$; the face part of the frame was about $100\times 100$. The database includes meshes of 1204 distinct Caucasian individuals, ages 3-40 obtained by a 3dMD digital stereophotogrammetry system. The database does not include texture or color information due to privacy. Each mesh includes 15K-20K vertices. Subjects all face forward, have a neutral expression, and wear caps to remove  hair occlusions. Meshes are cleaned by trained personnel and 15 anatomical facial landmarks were manually labeled by a single trained expert. Figure \ref{fig:database} shows examples of 3D meshes produced by the 3dMD system. The landmarks are used to register all the meshes to each other using \cite{allen}. 

The experiments were run on an Intel Xeon 2.67GHz/2.66GHz CPU, 16GB RAM in Windows Server 2008 R2 64bit environment. For a typical result mesh of 15K vertices, the running time was $92.16$s, with $1.2$s for preprocessing (finding fiducial points, rigid alignment), $83.4$s for non-rigid registration, $7.16$s for retrieval (calculating features for the input, warping all the faces, finding the best matching parts), and $0.4$s for merging. The non-rigid registration part ($90\%$ of the running time) could be replaced with a real-time registration method~\cite{zollhofer2014real, kazemi2014real}.  

\subsection{Evaluation of similarity function}\label{sec:sim_evaluation}
\begin{figure}[ht]
  \centering
  \includegraphics[width=.93\linewidth]{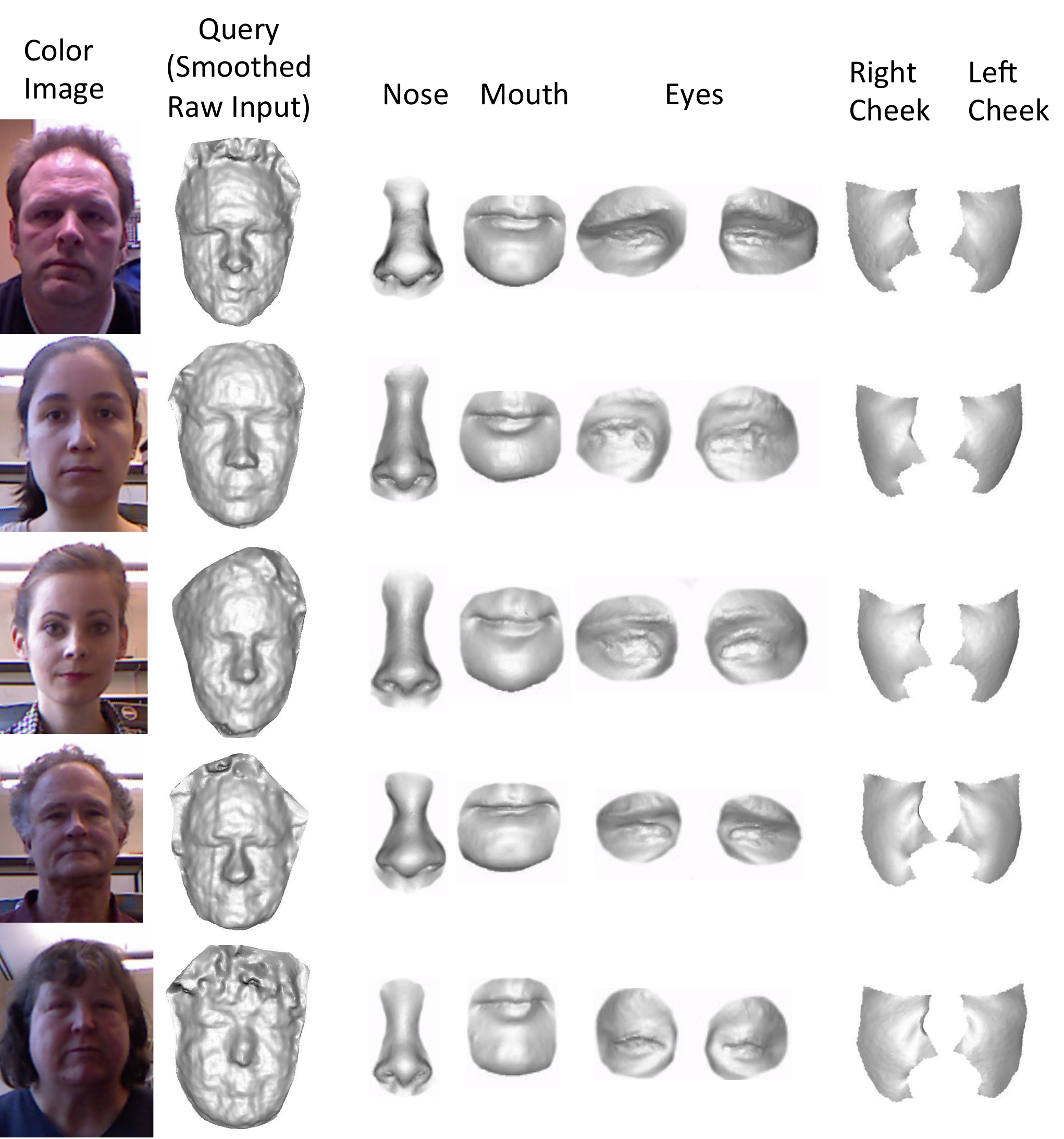}
  \caption{Similar parts that were retrieved using our approach. Photo shown only for reference. }\label{similarparts}
\end{figure}

To evaluate our similarity measure we tested it with seven ground-truth meshes ($S1-S7$). We included the ground-truth meshes in the database, and retrieved the best mesh per facial part. The inputs were Kinect depth images of the corresponding people. We compared pseudo-landmarks and azimuth-elevation histogram contributions at different resolutions as well as our final combined similarity distance. For each person, we obtained the ranking of the ground-truth in the retrieval results (lower is better). Note that the ground-truth meshes and Kinect inputs are not exactly the same, since the facial expression of the person may slightly change between the two captures. Tables \ref{similaritynose}, \ref{similaritycheek}, \ref{similaritymouth}, and \ref{similarityeyes}  show the rankings for nose, cheeks, mouth and eyes areas respectively. Most of the cases show that increasing the resolution of pseudo-landmarks does not improve the retrieval result. As shown in Tables \ref{similaritynose} and \ref{similaritycheek}, the similarity
function using the combined features worked extremely well on retrieving
based on similarity of the nose and cheeks. For the nose, two individuals
were returned as best matches, two others as second best, and another as
third best (out of 1204 + 7 = 1211). For the cheeks, the similarity
function with combined features returned the correct individuals with
rankings of five through 68. The mouth region proved to be a little more
difficult with the correct individuals achieving rankings from 1 to 229.
The eyes were the most difficult with rankings from 12 to 482. We note that
the eyes are the worst part of the Kinect depth frames, often not showing up
well at all. 
 Most of the obtained rankings were in the top 10\% of the 1211
possible individuals in the expanded database.
We show the five similar parts for input examples in Fig. \ref{similarparts}.
Note that while matching of 3D meshes is a widely studied research area \cite{kalogerakis2010learning,berretti20103d}, there is no prior work on matching a noisy depth frame to high resolution meshes. 

\begin{table}[bp]
\caption{Ranking from our distance function on the nose region.}
\label{similaritynose}
\begin{center}
\begin{tabular}{cccccccc}
\multicolumn{1}{c}{\bf{Dist.}} &\multicolumn{1}{c}{S1}  &\multicolumn{1}{c}{S2} &\multicolumn{1}{c}{S3} &\multicolumn{1}{c}{S4}  &\multicolumn{1}{c}{S5} &\multicolumn{1}{c}{S6}   
&\multicolumn{1}{c}{S7} 
\\ \hline \\
Pts 35x35 &157 &2 &809 &1 &14 &1 &58\\
Pts 65x65 &157 &2 &813 &1 &14 &1 &38\\
A-E hist    &24  &7 &1 &33 &99 &238 &9\\
Combined  &\bf{14}  &\bf{1} &\bf{3} &\bf{2} &\bf{14} &\bf{1} &\bf{2}\\
\end{tabular}
\end{center}
\end{table}

\begin{table}[bp]
\caption{Ranking from our distance function on the cheek region.}
\label{similaritycheek}
\begin{center}
\begin{tabular}{cccccccc}
\multicolumn{1}{c}{\bf{Dist.}} &\multicolumn{1}{c}{S1}  &\multicolumn{1}{c}{S2} &\multicolumn{1}{c}{S3} &\multicolumn{1}{c}{S4}  &\multicolumn{1}{c}{S5} &\multicolumn{1}{c}{S6}   
&\multicolumn{1}{c}{S7} 
\\ \hline \\
Pts 35x35 &17 &64 &88 &\bf{64} &49 &3 &89\\
Pts 65x65 &17 &76 &83 &70 &47 &3 &83\\
A-E hist    &229  &98 &47 &314 &334 &11 &38\\
Combined  &\bf{12} &\bf{16} &\bf{6} &68 &\bf{22} &\bf{5} &\bf{31}\\
\end{tabular}
\end{center}
\end{table}

\begin{table}[bp]
\caption{Ranking from our distance function on the mouth region.}
\label{similaritymouth}
\begin{center}
\begin{tabular}{cccccccc}
\multicolumn{1}{c}{\bf{Dist.}} &\multicolumn{1}{c}{S1}  &\multicolumn{1}{c}{S2} &\multicolumn{1}{c}{S3} &\multicolumn{1}{c}{S4}  &\multicolumn{1}{c}{S5} &\multicolumn{1}{c}{S6}   
&\multicolumn{1}{c}{S7} 
\\ \hline \\
Pts 35x35 &229 &408 &441 &73 &22 &619 &342\\
Pts 65x65 &227 &382 &478 &90 &22 &581 &276\\
A-E hist    &27  &108 &1 &119 &17 &95 &262\\
Combined &\bf{20}  &\bf{94} &\bf{1} &\bf{60} &\bf{2} &\bf{83} &\bf{229}\\
\end{tabular}
\end{center}
\end{table}

\begin{table}[bp]
\caption{Ranking from our distance function on the eyes region.}
\label{similarityeyes}
\begin{center}
\begin{tabular}{cccccccc}
\multicolumn{1}{c}{\bf{Dist.}} &\multicolumn{1}{c}{S1}  &\multicolumn{1}{c}{S2} &\multicolumn{1}{c}{S3} &\multicolumn{1}{c}{S4}  &\multicolumn{1}{c}{S5} &\multicolumn{1}{c}{S6}   
&\multicolumn{1}{c}{S7} 
\\ \hline \\
Pts 35x35 &92 &57 &543 &43 &102 &351 &475\\
Pts 65x65 &90 &67 &544 &56 &103 &395 &429\\
A-E hist    &184  &617 &484 &713 &334 &11 &231\\
Combined &\bf{47} &\bf{226} &\bf{482} &\bf{210} &\bf{75} &\bf{12} &\bf{75}\\
\end{tabular}
\end{center}
\end{table}

\subsection{Comparisons of reconstructions}
\begin{figure}
  \centering
  \includegraphics[width=0.48\textwidth]{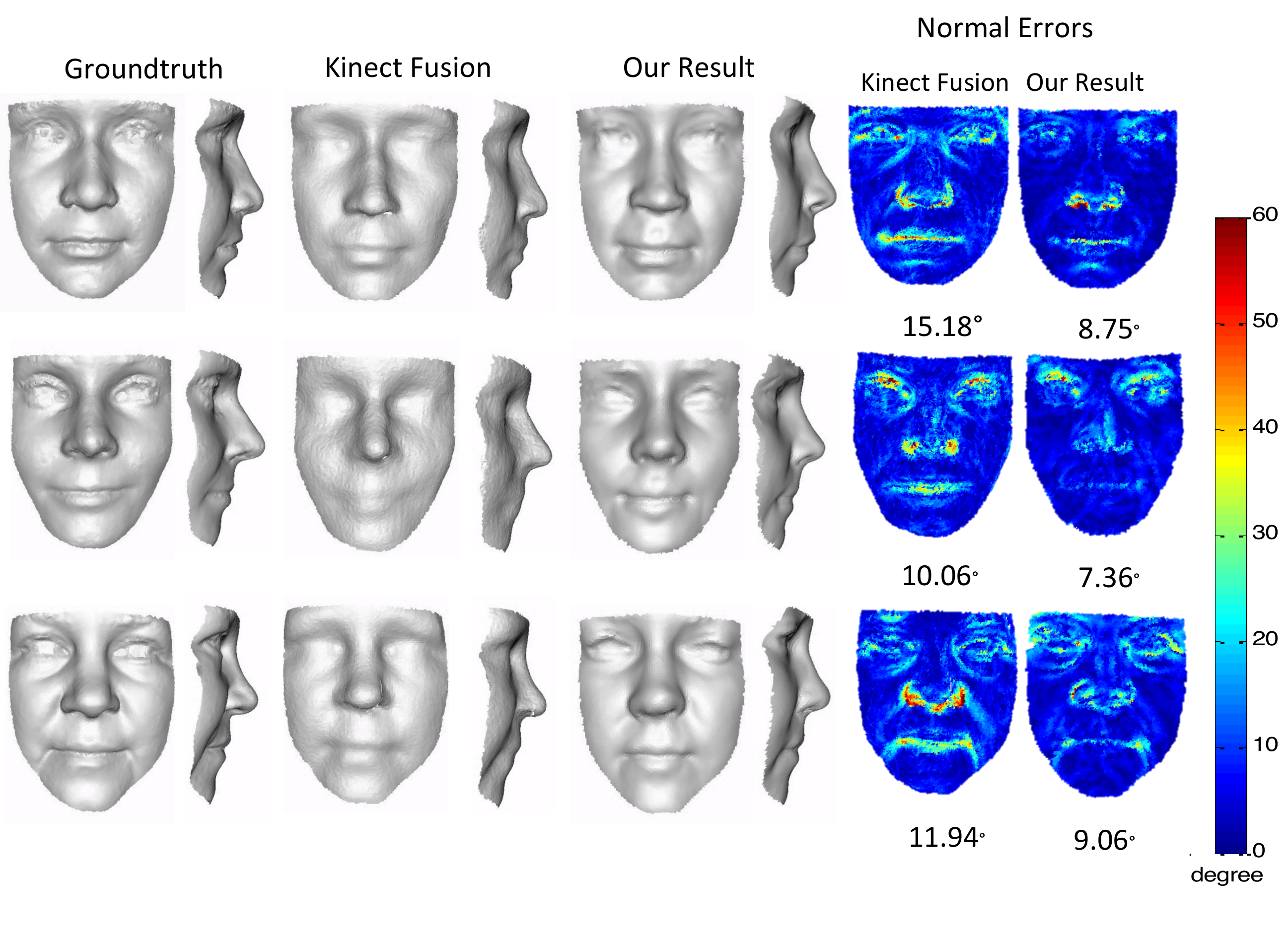}
  \caption{Comparison to ground-truth and KinectFusion~\cite{newcombe2011kinectfusion}.}\label{reconstructionerror}
\end{figure}

\begin{figure}
  \centering
  \includegraphics[width=.45\textwidth]{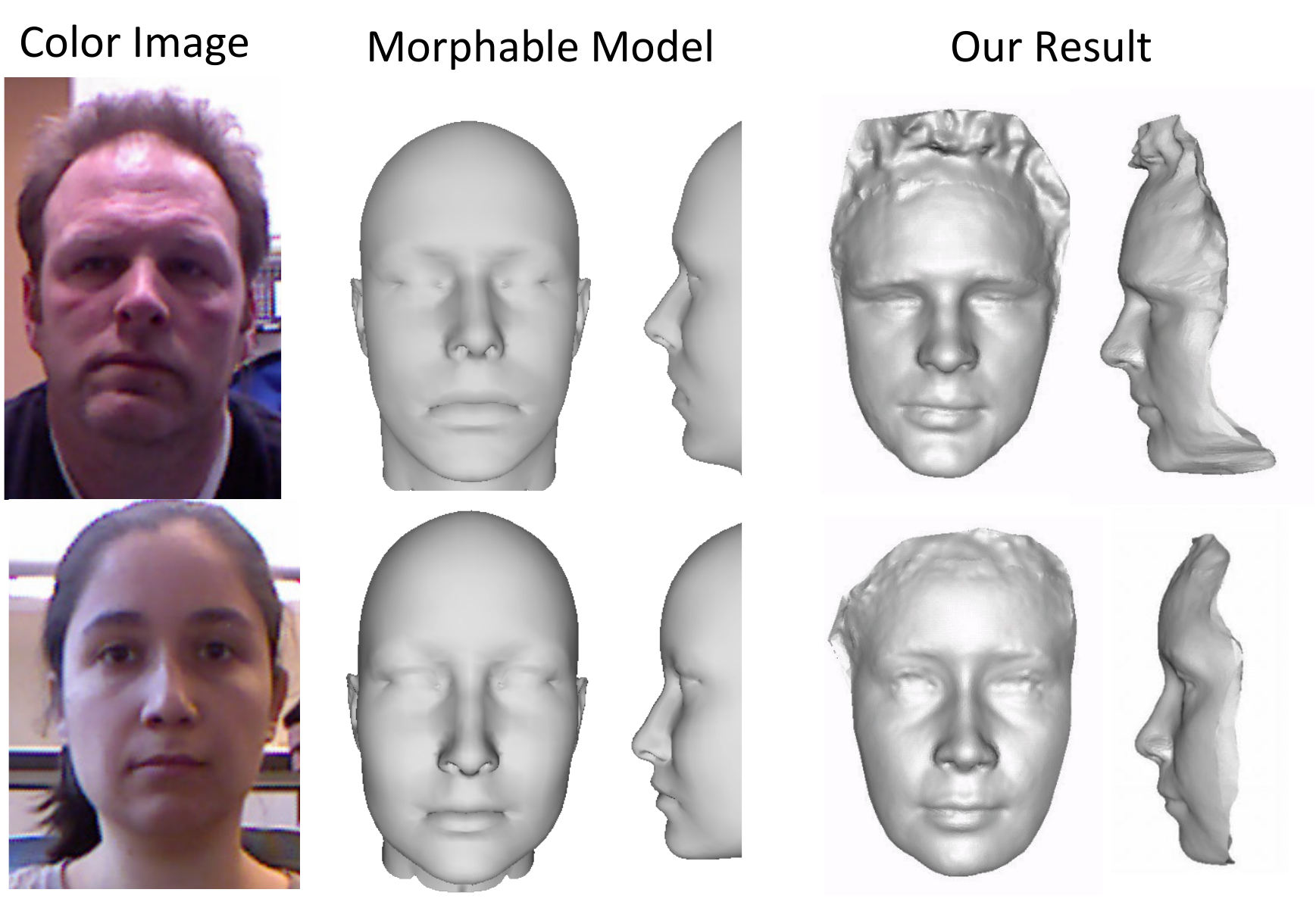}
  \caption{Comparison to reconstructions by Vizago (implementation of the morphable model approach) \cite{vizago}.}\label{comparisonMM}
\end{figure}

\begin{figure}
  \centering
  \includegraphics[width=.48\textwidth]{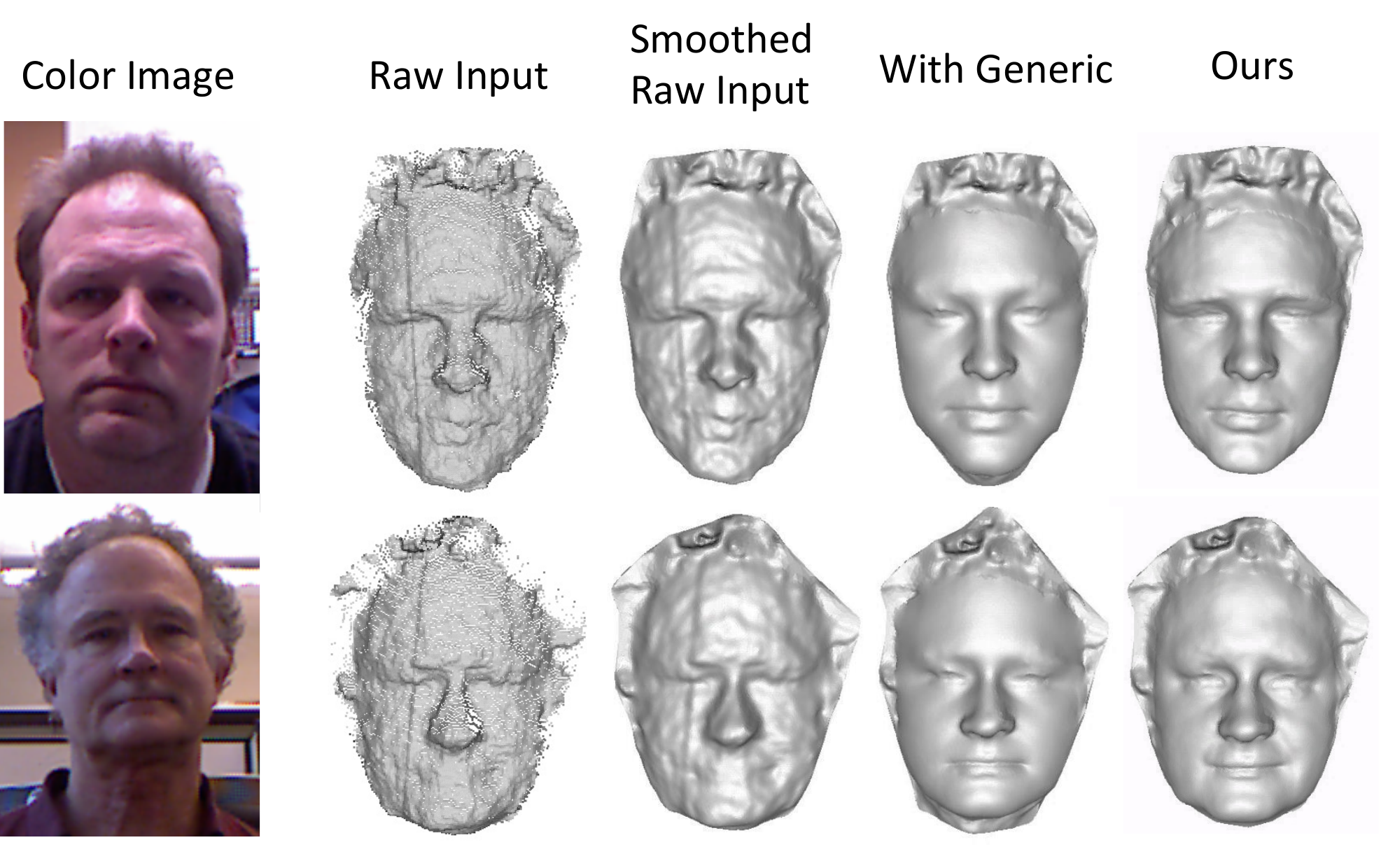}
  \caption{Comparison to smoothed Kinect frame and details from generic shape. Our method using matched facial parts, produces high-detail reconstruction than using a generic shape. }\label{comparison}
\end{figure}

\begin{figure}
  \centering
  \includegraphics[width=.4\textwidth]{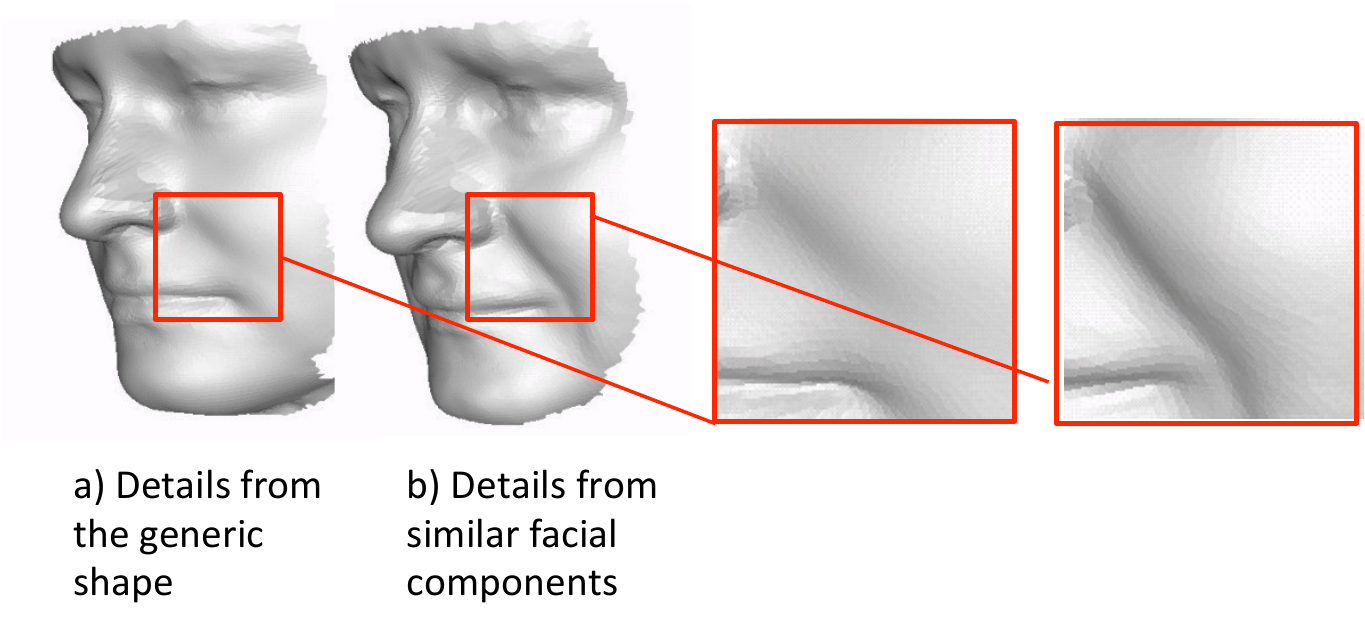}
  \caption{When a single generic shape (rather than the database) is used to  fill in high-resolution details, individual details are not captured. See also Fig.~\ref{comparison}.}\label{comparemean}
\end{figure}

We compared our reconstructions to reconstructions by KinectFusion \cite{newcombe2011kinectfusion}(implementation by Kinect for Windows SDK v1.8 \cite{KinectSDK}) and to ground-truth shapes for people who were not part of the original database (since the people in the original database are unknown IRB-protected subjects). KinectFusion requires the subject to stay still and requires a few dozen Kinect frames, while our method requires a single frame. For each reconstruction we show the meshes and the error in surface normals (in angles). Fig.~\ref{reconstructionerror} shows the results on three meshes from our test set and includes the angle error for both KinectFusion and our result. In all tests, our result had a lower error than KinectFusion. We next compared our results to those generated using a morphable model technique (online implementation by Vizago \cite{vizago}). Fig. \ref{comparisonMM} shows that the morphable model results are very dependent on their database and produce somewhat generic results, while our results capture more individual details. We have also tested the contribution of using the database vs. just using the generic shape and non-rigid registration for the reconstruction and filling in the missing details in Kinect depth as shown in Fig. \ref{comparison}. Note that facial details are not captured with the generic model but appear once the database is used, as shown in Fig.~\ref{comparemean}.

\subsection{Additional results}
\begin{figure*}
  \centering
  \includegraphics[width=.99\textwidth]{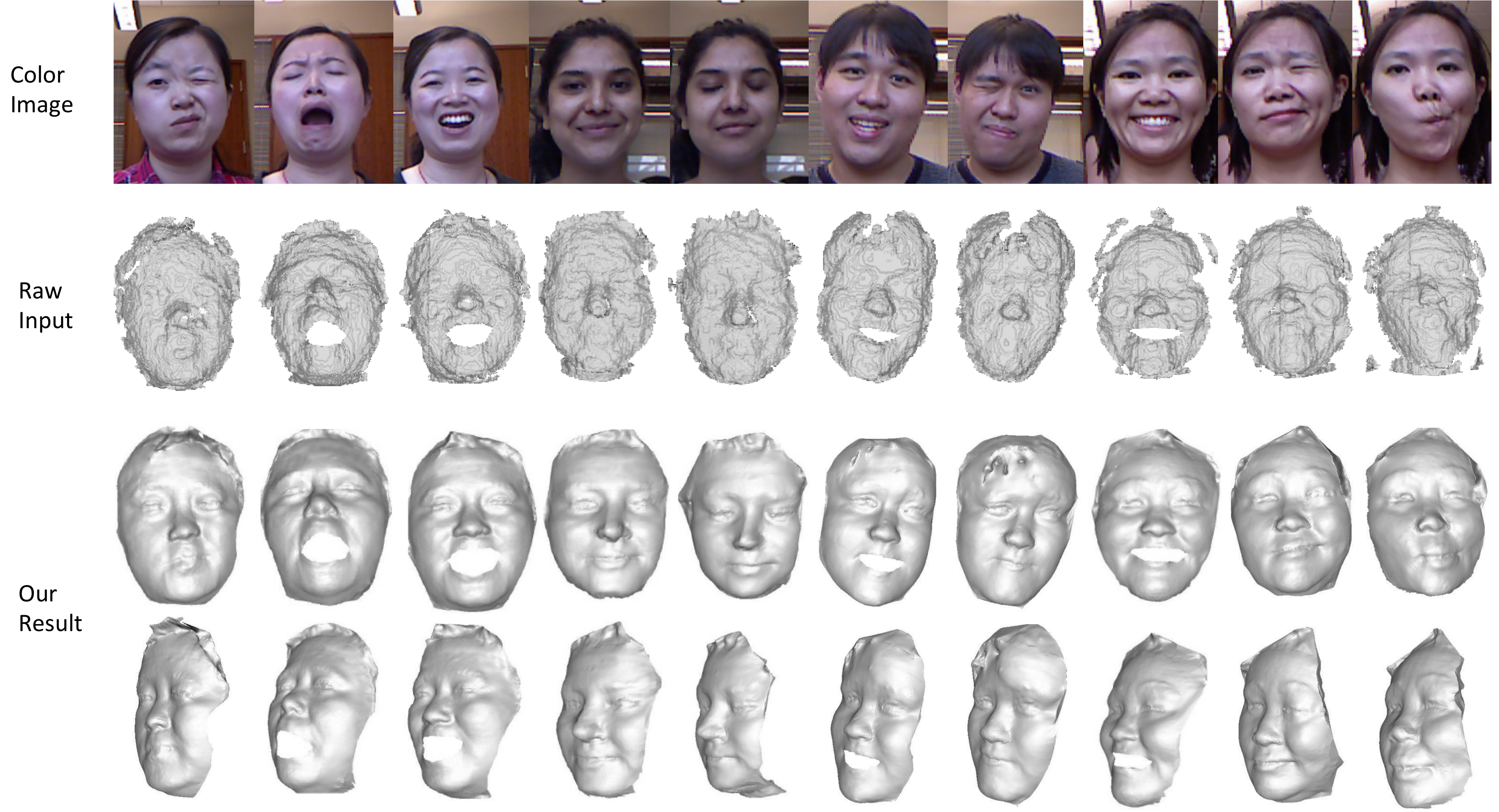}
  \caption{Reconstructions of facial expressions. }\label{expression}
\end{figure*}

\begin{figure*}
  \centering
  \includegraphics[height=0.95\textheight]{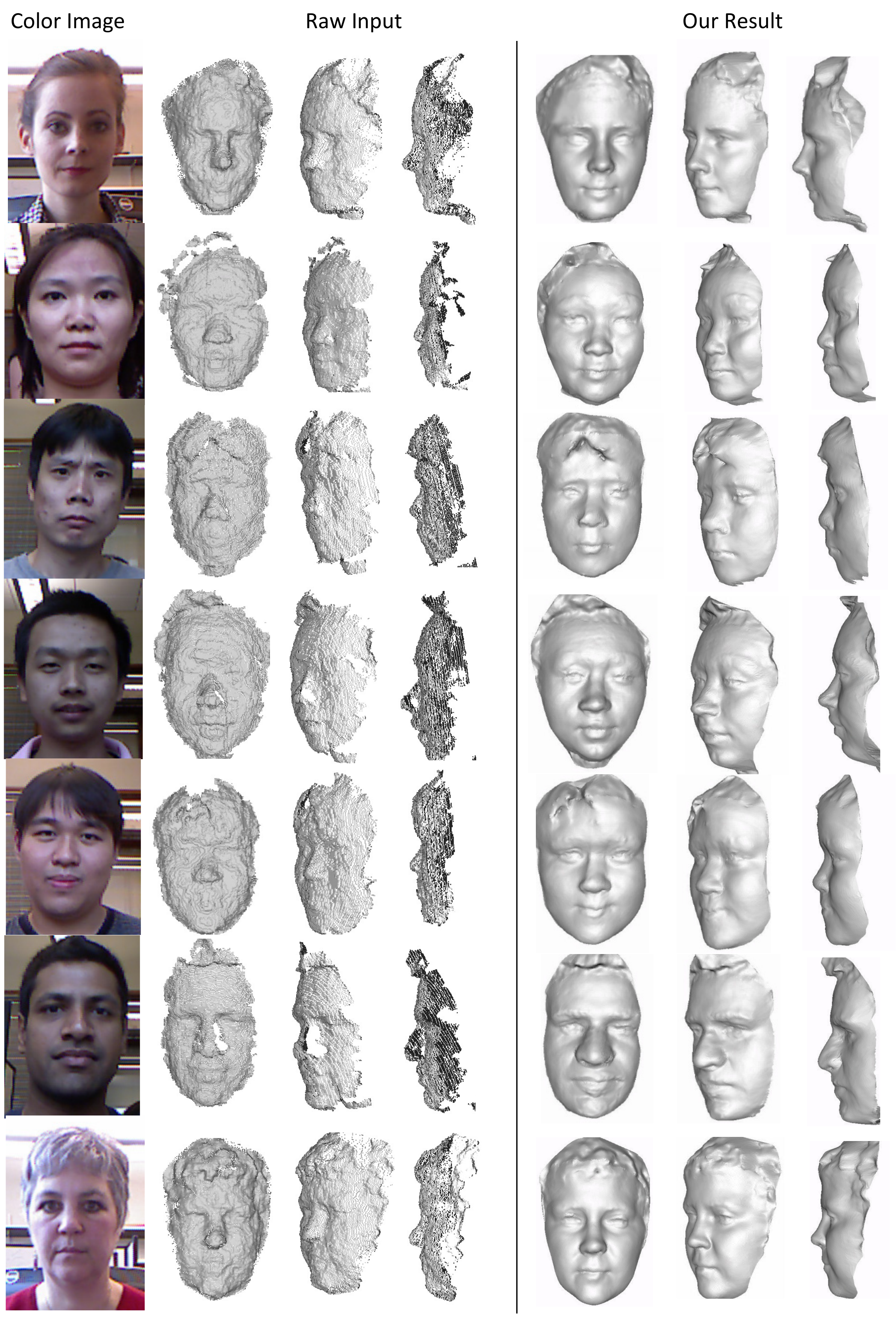}
  \caption{Additional results. Observe how well the shape is reconstructed, even though some of the people are of different ethnicity (the database includes only Caucasians) and age. The RGB image is provided only for reference and is not used in the matching. The input is a \textbf{single} Kinect  frame (shown from different sides). Note that Asians and people over 40 are not part of the database. }\label{otherresult}
\end{figure*}

Fig. \ref{expression} shows reconstructions of facial expressions from a single Kinect frame (given a neutral face frame). Fig.~\ref{otherresult} shows additional results, most of which did not have a ground-truth mesh. However, it is interesting to observe that the facial shape is reconstructed very well even though some of the people are not in the age span of the database or have a different ethnicity. The method is invariant to imaging conditions (light, pose) since the reconstruction is done based on depth-to-mesh matching and does not use the color channels.

%% file: conclusion.tex
\section{Conclusion}\label{sec:conclusion}
In this paper, we described our approach for reconstruction of a high-quality 3D face mesh from a rough, noisy, low-resolution single Kinect depth frame. We leveraged a large dataset of high-resolution meshes of distinct individuals. Within that method, we have defined and tested a similarity measure that uses a linear combination of pseudo-landmark points and an azimuth-elevation angle histogram to retrieve parts of dataset faces that are most similar to the semantically equivalent parts of the query face. Our key contribution is to show that extremely simple part-based matching to a large set of faces enables the creation of remarkably accurate high-resolution meshes of novel people from  noisy single-frame input. The resultant meshes can be further used for facial expression modeling, as we also demonstrated.

%% file: egpaper_final.bbl
\begin{thebibliography}{10}\itemsep=-1pt

\bibitem{alexander2013digital}
O.~Alexander, G.~Fyffe, J.~Busch, X.~Yu, R.~Ichikari, A.~Jones, P.~Debevec,
  J.~Jimenez, E.~Danvoye, B.~Antionazzi, et~al.
\newblock Digital ira: creating a real-time photoreal digital actor.
\newblock In {\em ACM SIGGRAPH 2013 Posters}, page~1. ACM, 2013.

\bibitem{alexander2009digital}
O.~Alexander, M.~Rogers, W.~Lambeth, M.~Chiang, and P.~Debevec.
\newblock The digital emily project: photoreal facial modeling and animation.
\newblock In {\em ACM SIGGRAPH 2009 Courses}, page~12. ACM, 2009.

\bibitem{allen}
B.~Allen, B.~Curless, and Z.~Popovi{\'c}.
\newblock The space of human body shapes: reconstruction and parameterization
  from range scans.
\newblock In {\em ACM Transactions on Graphics (TOG)}, volume~22, pages
  587--594. ACM, 2003.

\bibitem{anguelov2005scape}
D.~Anguelov, P.~Srinivasan, D.~Koller, S.~Thrun, J.~Rodgers, and J.~Davis.
\newblock Scape: shape completion and animation of people.
\newblock In {\em ACM Transactions on Graphics (TOG)}, volume~24, pages
  408--416. ACM, 2005.

\bibitem{atmosukarto2010use}
I.~Atmosukarto, L.~G. Shapiro, and C.~Heike.
\newblock The use of genetic programming for learning 3d craniofacial shape
  quantifications.
\newblock In {\em Pattern Recognition (ICPR), 2010 20th International
  Conference on}, pages 2444--2447. IEEE, 2010.

\bibitem{beeler2010high}
T.~Beeler, B.~Bickel, P.~Beardsley, B.~Sumner, and M.~Gross.
\newblock High-quality single-shot capture of facial geometry.
\newblock {\em ACM Transactions on Graphics (TOG)}, 29(4):40, 2010.

\bibitem{beeler2011high}
T.~Beeler, F.~Hahn, D.~Bradley, B.~Bickel, P.~Beardsley, C.~Gotsman, R.~W.
  Sumner, and M.~Gross.
\newblock High-quality passive facial performance capture using anchor frames.
\newblock In {\em ACM Transactions on Graphics (TOG)}, volume~30, page~75. ACM,
  2011.

\bibitem{berretti20103d}
S.~Berretti, A.~Del~Bimbo, and P.~Pala.
\newblock 3d face recognition using isogeodesic stripes.
\newblock {\em Pattern Analysis and Machine Intelligence, IEEE Transactions
  on}, 32(12):2162--2177, 2010.

\bibitem{blanz1999morphable}
V.~Blanz and T.~Vetter.
\newblock A morphable model for the synthesis of 3d faces.
\newblock In {\em Proceedings of the 26th annual conference on Computer
  graphics and interactive techniques}, pages 187--194. ACM
  Press/Addison-Wesley Publishing Co., 1999.

\bibitem{bouaziz2013online}
S.~Bouaziz, Y.~Wang, and M.~Pauly.
\newblock Online modeling for realtime facial animation.
\newblock {\em ACM Transactions on Graphics (TOG)}, 32(4):40, 2013.

\bibitem{bradley2010high}
D.~Bradley, W.~Heidrich, T.~Popa, and A.~Sheffer.
\newblock High resolution passive facial performance capture.
\newblock {\em ACM Transactions on Graphics (TOG)}, 29(4):41, 2010.

\bibitem{cao2014face}
X.~Cao, Y.~Wei, F.~Wen, and J.~Sun.
\newblock Face alignment by explicit shape regression.
\newblock {\em International Journal of Computer Vision}, 107(2):177--190,
  2014.

\bibitem{desbrun1999implicit}
M.~Desbrun, M.~Meyer, P.~Schr{\"o}der, and A.~H. Barr.
\newblock Implicit fairing of irregular meshes using diffusion and curvature
  flow.
\newblock In {\em Proceedings of the 26th annual conference on Computer
  graphics and interactive techniques}, pages 317--324. ACM
  Press/Addison-Wesley Publishing Co., 1999.

\bibitem{fan2014learning}
H.~Fan, Z.~Cao, Y.~Jiang, Q.~Yin, and C.~Doudou.
\newblock Learning deep face representation.
\newblock {\em arXiv preprint arXiv:1403.2802}, 2014.

\bibitem{fanelli2013random}
G.~Fanelli, M.~Dantone, J.~Gall, A.~Fossati, and L.~Van~Gool.
\newblock Random forests for real time 3d face analysis.
\newblock {\em International Journal of Computer Vision}, 101(3):437--458,
  2013.

\bibitem{ghosh2011multiview}
A.~Ghosh, G.~Fyffe, B.~Tunwattanapong, J.~Busch, X.~Yu, and P.~Debevec.
\newblock Multiview face capture using polarized spherical gradient
  illumination.
\newblock {\em ACM Transactions on Graphics (TOG)}, 30(6):129, 2011.

\bibitem{gower1975generalized}
J.~C. Gower.
\newblock Generalized procrustes analysis.
\newblock {\em Psychometrika}, 40(1):33--51, 1975.

\bibitem{vizago}
G.~R. Group.
\newblock Vizago.
\newblock \url{http://www.vizago.ch}.

\bibitem{hays2007scene}
J.~Hays and A.~A. Efros.
\newblock Scene completion using millions of photographs.
\newblock In {\em ACM Transactions on Graphics (TOG)}, volume~26, page~4. ACM,
  2007.

\bibitem{face++}
M.~Inc.
\newblock Face++ research toolkit.
\newblock \url{www.faceplusplus.com}, Dec. 2013.

\bibitem{kalogerakis2010learning}
E.~Kalogerakis, A.~Hertzmann, and K.~Singh.
\newblock Learning 3d mesh segmentation and labeling.
\newblock {\em ACM Transactions on Graphics (TOG)}, 29(4):102, 2010.

\bibitem{kazemi2014real}
V.~Kazemi, C.~Keskin, T.~Jonathan, K.~Pushmeet, and I.~Shahram.
\newblock Real-time face reconstruction from a single depth image.
\newblock 2014.

\bibitem{kemelmacher20113d}
I.~Kemelmacher-Shlizerman and R.~Basri.
\newblock 3d face reconstruction from a single image using a single reference
  face shape.
\newblock {\em Pattern Analysis and Machine Intelligence, IEEE Transactions
  on}, 33(2):394--405, 2011.

\bibitem{kovnatsky2013coupled}
A.~Kovnatsky, M.~M. Bronstein, A.~M. Bronstein, K.~Glashoff, and R.~Kimmel.
\newblock Coupled quasi-harmonic bases.
\newblock In {\em Computer Graphics Forum}, volume~32, pages 439--448. Wiley
  Online Library, 2013.

\bibitem{mercan2013use}
E.~Mercan, L.~G. Shapiro, S.~M. Weinberg, and S.-I. Lee.
\newblock The use of pseudo-landmarks for craniofacial analysis: A comparative
  study with l 1-regularized logistic regression.
\newblock In {\em Engineering in Medicine and Biology Society (EMBC), 2013 35th
  Annual International Conference of the IEEE}, pages 6083--6086. IEEE, 2013.

\bibitem{KinectSDK}
Microsoft.
\newblock Kinect for windows software development kit v1.8.
\newblock \url{http://www.microsoft.com/en-us/kinectforwindows/}.

\bibitem{nehab2005efficiently}
D.~Nehab, S.~Rusinkiewicz, J.~Davis, and R.~Ramamoorthi.
\newblock Efficiently combining positions and normals for precise 3d geometry.
\newblock In {\em ACM Transactions on Graphics (TOG)}, volume~24, pages
  536--543. ACM, 2005.

\bibitem{newcombe2011kinectfusion}
R.~A. Newcombe, A.~J. Davison, S.~Izadi, P.~Kohli, O.~Hilliges, J.~Shotton,
  D.~Molyneaux, S.~Hodges, D.~Kim, and A.~Fitzgibbon.
\newblock Kinectfusion: Real-time dense surface mapping and tracking.
\newblock In {\em Mixed and augmented reality (ISMAR), 2011 10th IEEE
  international symposium on}, pages 127--136. IEEE, 2011.

\bibitem{pighin2006synthesizing}
F.~Pighin, J.~Hecker, D.~Lischinski, R.~Szeliski, and D.~H. Salesin.
\newblock Synthesizing realistic facial expressions from photographs.
\newblock In {\em ACM SIGGRAPH 2006 Courses}, page~19. ACM, 2006.

\bibitem{pokrass2013partial}
J.~Pokrass, A.~M. Bronstein, and M.~M. Bronstein.
\newblock Partial shape matching without point-wise correspondence.
\newblock {\em Numerical Mathematics: Theory, Methods \& Applications}, 6(1),
  2013.

\bibitem{weise2011realtime}
T.~Weise, S.~Bouaziz, H.~Li, and M.~Pauly.
\newblock Realtime performance-based facial animation.
\newblock {\em ACM Transactions on Graphics (TOG)}, 30(4):77, 2011.

\bibitem{zhang2007spacetime}
L.~Zhang, N.~Snavely, B.~Curless, and S.~M. Seitz.
\newblock Spacetime faces: High-resolution capture for\~{} modeling and
  animation.
\newblock In {\em Data-Driven 3D Facial Animation}, pages 248--276. Springer,
  2007.

\bibitem{zollhofer2014real}
M.~Zollh{\"o}fer, M.~Nie{\ss}ner, S.~Izadi, C.~Rehmann, C.~Zach, M.~Fisher,
  C.~Wu, A.~Fitzgibbon, C.~Loop, C.~Theobalt, et~al.
\newblock Real-time non-rigid reconstruction using an rgb-d camera.
\newblock {\em ACM Transactions on Graphics, TOG}, 2014.

\end{thebibliography}
